\documentclass[11pt,a4paper]{article}
\usepackage[hyperref]{naaclhlt2019}
\usepackage{times}
\usepackage{tipa}
\usepackage{latexsym}
\usepackage{booktabs}
\usepackage{graphicx}
\usepackage{tikz}
\usepackage{pgfplots}
\usepackage{url}

\aclfinalcopy % Uncomment this line for the final submission
\title{Massively Multilingual Adversarial Speech Recognition}

\author{Oliver Adams, Matthew Wiesner, Shinji Watanabe, and David Yarowsky \\
  Department of Computer Science \\
  Johns Hopkins University, Baltimore MD, USA \\
  {\tt \{oadams1,wiesner,shinjiw,yarowsky\}@jhu.edu}}

\date{}

\begin{document}
\maketitle
\begin{abstract}
We report on adaptation of multilingual end-to-end speech recognition models trained on as many as 100 languages. Our findings shed light on the relative importance of similarity between the target and pretraining languages along the dimensions of phonetics, phonology, language family, geographical location, and orthography. In this context, experiments demonstrate the effectiveness of two additional pretraining objectives in encouraging language-independent encoder representations: a context-independent phoneme objective paired with a language-adversarial classification objective.
\end{abstract}

\section{Introduction}

The main difficulty in creating automatic speech recognition (ASR) systems for
a large number of the world's 7,000 languages is a lack of training data. Such
data comes in the form of speech paired with transcriptions, a pronunciation
lexicon, and text for language model training. A common technique in data-constrained settings is to learn language-independent representations of speech via multilingual training. Popular
approaches include the use of multilingual bottleneck features
\cite{vesely2012language} as well as multilingual model training before
fine-tuning to a \emph{target} language \cite{scanzio2008use,vu2012multilingual}.

Prior work in multilingual and cross-lingual speech recognition has been
restricted to a small handful of the world's most-spoken languages, relying on
multilingual corpora such as GlobalPhone \cite{schultz2002globalphone}, the
IARPA Babel corpora \cite{gales2014speech}, or the VoxForge\footnote{\url{voxforge.org}} corpora. Most work
typically only reports on models trained on a subset of these languages.

%In this paper, we report on the application of multilingual ASR to as many as 100 languages using Bible data that comprises the CMU Wilderness dataset.
In this paper we explore pretraining multilingual ASR models using speech from as many as 100 languages from the CMU Wilderness Multilingual Speech Dataset \cite{black2019cmu}.\footnote{\url{festvox.org/cmu\_wilderness/index.html}} To the best of our knowledge, this is the greatest number of  languages that has been
used in multilingual ASR model training to date.
We perform experiments to guide the choice of languages used when pretraining the model and assess the relative importance of similarity between the pretraining languages and target language in terms of geographic location, phonology, phonetic inventory, language family and orthography. 

We examine these variables in the context of two experimental setups: one where models are adapted to target language and target speakers, and one where models are adapted to target language but non-target speakers. The first task is relevant to language documentation contexts, which often involves transcribing speech of specific speakers for which there already exists some transcribed speech as training data \cite{michaud2018integrating}. The second case is relevant to incident response as modelled by LORELEI \cite{Strassel2016}, where there may only be a single target-language consultant available for which transcribed speech can be elicited, but the goal is to have an ASR model that generalizes to multiple speakers. 

Multilingual ASR training on such a scale presents challenges because of this language diversity.
In order to guide the model to learn language-independent representations
that are more amenable to adaptation, we experiment with two auxiliary training
tasks. The first is context-independent phoneme sequence prediction
to help bridge orthographic inconsistencies between languages. The
second is a domain-adversarial classification objective \cite{ganin2016domain}
over languages to encourage invariance of the model with respect to
language-specific phenomena. The hierarchical combination of grapheme and phoneme objectives
has only been used in monolingual end-to-end frameworks
\cite{krishna2018hierarchical,Rao2017}. Language-adversarial training in ASR \cite{Yi2018} has not been done at this scale before, nor in an end-to-end framework.

Our experiments are designed to answer the following questions:

\begin{enumerate}
	\item Is there benefit in scaling multilingual model training to a large
	number of languages?
	\item In what circumstances, if any, does the addition of a phoneme and/or
	language-adversarial objective improve multilingual models?
	\item How should we choose languages with which to pretrain a multilingual
	model?
	\item Do the answers to the above questions change when adapting to
	target versus non-target speakers in the target language?
\end{enumerate}

We find that using the auxiliary objectives in pretraining facilitates model transfer to unseen languages, especially when the pretraining languages are very dissimilar (Section \ref{sec:pilot}).
When the target speakers are seen in adaptation (Section \ref{sec:lang_sim}), similarity of the pretraining languages and the target language is
more important than quantity of pretraining languages. Choosing as pretraining languages geographically proximal languages tends
to help more than phonetically and phonologically similar but otherwise distant 
languages. However, when adapting to a handful of non-target speakers of the target language (Section \ref{sec:notgt}), the domain mismatch caused by the unseen speaker, language, or recording environment degrades performance. Exposing the model to as many pretraining languages as possible becomes vital to minimize this mismatch. Results on this task demonstrate that a massively multilingual seed model substantially outperforms other seed models trained on languages similar to the target. We will provide an ESPnet recipe to train and test our models.

%\begin{itemize}
%	\item If you had to pick languages, go with geographically similar ones.
%	Phon/Inv ones are good but not if the languages vary in other dimensions.
%	\item for tgt adaptation, more languages isn't necessarily better.
%	\item For notgt adaptation more languages is better and the other factors
%	don't matter as much.
%	\item Similarity of phonetic inventory doesn't seem to be as useful as
%	phonological + phonetic similarity.
%\end{itemize}

%Perhaps it's worth noting that we're not so concerned about producing a state-of-the-art ASR model, but rather exploring the value of different choices of pretrained languages and how best to work them together.

\section{Related Work}

This paper builds on work on multilingual ASR, end-to-end ASR, and adversarial learning.

\paragraph{Multilingual transfer} in ASR often relies on using bottle-neck features \cite{vesely2012language,vu2012multilingual,karafiat2018analysis} and adapting an acoustic model trained on one language to effectively recognize the sounds of other languages \cite{Schultz2001,le2005first,stolcke2006cross,toth2008cross,plahl2011cross,thomas2012multilingual,imseng2014using,do2014cross, heigold2013multilingual,scharenborg2017building}. However, while most work uses less than 10 languages for model training, we include up to 100 languages in training.
\paragraph{End-to-end ASR} has recently become popular, with approaches such as attention-based encoder-decoder models \cite{Chorowski2015,chan2015listen}, the connectionist temporal classification (CTC) objective of \newcite{Graves2006,graves2013speech}, or a combination of both \cite{kim2016joint,hori2017advances}. These approaches have also been deployed in multilingual settings \cite{toshniwal2017multilingual,chiu2017state,mueller2017phonemic,dalmia2018sequence,watanabe2017language}.
Our baseline approach to multilingual knowledge transfer is most similar to \newcite{inaguma2018transfer}, and involves training a hybrid CTC-attention seed model.

\paragraph{Hierarchical and multi-task} approaches including combining grapheme and phoneme prediction in monolingual contexts \cite{Rao2017,krishna2018hierarchical} at different levels of the network, or using sub-word units of varying granularity \cite{sanabria2018hierarchical}, have been shown to improve ASR performance. In this paper we extend the approach of hierarchical placement of additional objectives in order to enforce language independent, transferable models.

%\paragraph{Domain-adversarial training} One contribution of our work is the use of a domain-adversarial classification objective \cite{ganin2016domain} over a large number of languages.
\paragraph{Domain-adversarial training} is one such method for encouraging the model to learn language independent representations. A key contribution of this paper is the use of a domain-adversarial classification objective \cite{ganin2016domain} over many languages in order to encourage the model to learn representations that are invariant to language. Domain-adversarial training incorporates an
auxiliary domain classification task, but negates gradients for
encoder weights before the parameter update in order to guide the encoder to produce hidden
representations that fool the classifier: i.e. they minimize information about
the language while still facilitating the primary task of speech recognition.

Domain-adversarial training has been used in speech recognition to learn
features invariant to noise conditions
\cite{Shinohara2016}, accents \cite{sun2017domain}, and sex \cite{tripathi2018adversarial}. Most closely related to our work is that of \newcite{Yi2018}, who use a language-adversarial objective when preparing multilingual bottleneck features from four languages for a hidden Markov model (HMM) ASR pipeline. In contrast, our work uses an adversarial objective across many languages, pairing it with a context-independent phoneme objective in an end-to-end framework.

\section{Data}

We scraped the data that forms the CMU Wilderness dataset, using a freely available
script.\footnote{\url{https://github.com/festvox/datasets-CMU\_Wilderness}} This dataset consists of dramatized readings of the Bible in hundreds of languages. Each reading is ascribed a rating based on alignment quality which fits into one of these classes: \texttt{very good}, \texttt{good}, \texttt{okay}, and \texttt{not okay}.

The script used to preprocess the data uses a universal pronunciation module
in Festival \cite{taylor1998architecture}\footnote{\url{http://www.cstr.ed.ac.uk/projects/festival/}} to
produce pronunciation lexicons using an approach based on that of UniTran \cite{yoon2007multilingual}, which we use to create phonemic transcriptions.
%\textbf{It in turn is based on the sort of
%pronunciations in unicode.  We have added a lot of different latin (and
%latin++) characters that weren't in the original but importantly we also
%predict zeron, one or more sampa phones grapheme, and never allow multiple
%graphemes to got ot one phone.''}). 

%These pronunciation lexicons have issues. For example in Portuguese all of the /k/ sounds got mapped to /ch/, while all of the /ch/ sounds got mapped to /t/. For monolingual systems this isn't problematic since its internally consistent, but in relating similar phonemes between languages, this can cause issues.

%An alternative would be to use the LanguageNet G2P rules,\footnote{\url{http://www.isle.illinois.edu/sst/data/g2ps/}} which would restrict us to the 130 or so languages on that list.

\subsection{Characteristics of the Speech} The dataset consists of readings of the Bible, with readings typically of just a few speakers, mostly male. These are often dramatized, with sound effects and background music. For many purposes this could be considered a limitation of the data. Although the characteristics of the speech are unique, it allows us to investigate multilingual models over many languages without the confounds of an overly noisy environment. It is not unreasonable to expect our findings to generalize to other speech recognition domains.

\subsection{Evaluation Languages}
While the dataset includes only a single reading of the Bible for most languages, there are a number with two or more. 
We evaluate on languages for which we can find two or more readings. This is so that we can compare adaptation to a target language but not the speakers of the target reading (we refer to this task as \emph{language adaptation}, as explored in Section \ref{sec:notgt}) with adaptation to the target language as well as the target reading (we refer to this task as \emph{reading adaptation}). We additionally restricted the evaluation languages to those that have at least one \texttt{good} or \texttt{very good} reading in terms of alignment quality. Table \ref{tab:eval_stats} presents the evaluation languages and readings grouped by family or geographic location, along with their durations.

 \begin{table}[]
    \centering
    \resizebox{\columnwidth}{!}{%
    \begin{tabular}{ccccc}
    \toprule
        & \multicolumn{3}{c}{Hours:minutes/quality per reading} \\
    \midrule
    Aymara (ayr) & 16:19/G & 18:37/G & - \\ %AYMBSB, AYMSBU
    SB Quechua (quh) & 27:41/G & 20:02/G & - \\ % QUHRBV, QUHSBB; both southern Bolivian.
    \midrule
    Kekchi (kek) & 19:32/G & 18:30/G & - \\ % KEKIBS, KEKSBG
%    Mazatec & 10:44 & 09:37 & - & O/O \\ %MAASAV, MAASJV
    Ixil (ixl) & 35:06/VG & 25:35/G & 18:29/G \\ %IX1WBT, IXIWBT, IXLWBT
   \midrule
    Malagasy (mlg) & 12:29/NO & 15:52/O & 15:59/G \\ %MLGEIV, MLGRCV, MLGRPV
    Indonesian (ind) & 19:01/G & 21:20/G & 30:34/G \\ % INZNTV, INZSHL, INZTSI
    \midrule
    Garap (kia) & 15:34/G & 12:17/VG & - \\ % KIABSC, KIAWBT
    \midrule
    Swedish (swe) & 15:55/G & 16:46/VG & - \\ % SWESFB, SWESFV
    Spanish (spn) & 16:35/G & 15:19/G & - \\ % SPNBDA, SPNWTC
    \bottomrule
    \end{tabular}
    }
    \caption{The duration of each reading in the evaluation languages (ISO 639-3 language codes in parentheses), before our preprocessing. Alignment quality categories are \texttt{very good} (VG), \texttt{good} (G), \texttt{okay} (O), \texttt{not okay} (NO). \emph{SB Quechua} denotes South Bolivian Quechua.}
    \label{tab:eval_stats}
    \vspace{-0.5cm}
\end{table}

\section{Auxiliary Training Objectives}
\label{sec:aux_objectives}

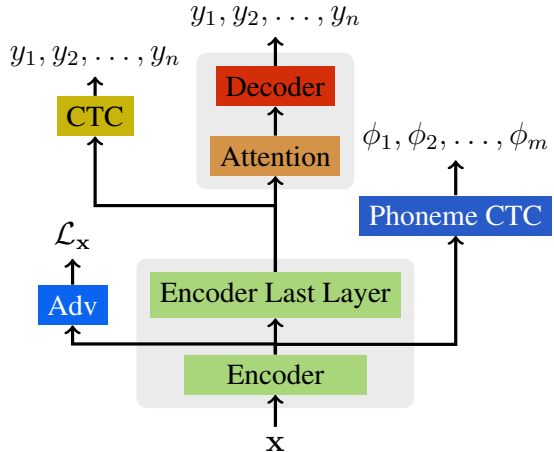
\begin{figure}
	\centering
\begin{tikzpicture}[x=0.75pt,y=0.75pt,yscale=-1,xscale=0.75]

\usetikzlibrary{fit}
\usetikzlibrary{intersections,shapes.arrows,calc}
\newcommand*{\connectorV}[4][]{
  \draw[#1] (#3) |- ($(#3) !#2! (#4)$) -| (#4);
}

\newcommand*{\connectorH}[4][]{
  \draw[#1] (#3) |- ($(#3) !#2! (#4)$) -| (#4);
}

% Input
\node (x) at (120,215) {\large $\mathbf{x}$} ;

% Encoder
\node[fill={rgb, 255:red, 184; green, 233; blue, 134 },rectangle,minimum width=2.4cm] (v1) at (120,180) {Encoder} ;
\draw[->,very thick] (x) to (v1) ;
\node[fill={rgb, 255:red, 184; green, 233; blue, 134 },rectangle,minimum width=2.4cm] (v2) at (120,140) {Encoder Last Layer} ;
\draw[->,very thick] (v1) to (v2) ;
\node[fill,opacity=0.08,rectangle,rounded corners,very thick,fit=(v1) (v2),inner sep=5pt] (rec) {} ;

% Decoder
\node[fill={rgb, 255:red, 230; green, 160; blue, 80 },rectangle] (v3) at (120,70) {Attention} ;
\node[fill={rgb, 255:red, 230; green, 50; blue, 10 },rectangle] (v4) at (120,35) {Decoder} ;
\draw[->,very thick] (v2) to (v3) ;
\draw[->,very thick] (v3) to (v4) ;
\node[fill,opacity=0.08,rectangle,rounded corners,very thick,fit=(v3) (v4),inner sep=5pt] (rec) {} ;

% Output
\node (y) at (120,0) {\large $y_1, y_2, \ldots, y_n$} ;
\draw[->,very thick] (v4) to (y) ;

\node (y2) at (0,20) {\large $y_1, y_2, \ldots, y_n$} ;

% CTC
\node[fill={rgb, 255:red, 200; green, 180; blue, 10 },rectangle] (v5) at (0,50) {CTC} ;
\connectorV[->,very thick]{0.5}{v2}{v5}
\draw[->,very thick] (v5) to (y2) ;

% Phoneme CTC
\node[fill={rgb, 255:red, 40; green, 90; blue, 200 },rectangle] (v6) at (240,100) {\textcolor{white}{Phoneme CTC}} ; 
\connectorV[->,very thick]{0.19}{v1}{v6}

% Phoneme output
\node (z) at (240,60) {\large $\phi_1, \phi_2, \ldots, \phi_m$} ;
\draw[->,very thick](v6) to (z) ;

% Language Adversarial Objective
\node[fill={rgb, 255:red, 10; green, 100; blue, 240 },rectangle] (v7) at (-15,145) {\textcolor{white}{Adv}} ;
\connectorV[->,very thick]{0.44}{v1}{v7}

% Language output
\node (l) at (-15,110) {\large $\mathcal{L_{\mathbf{x}}}$} ;
\draw[->,very thick] (v7) to (l) ;

\end{tikzpicture}
\caption{The end-to-end architecture used during pretraining. $\mathbf{x}$ is the input speech features, $y_1,y_2,\ldots,y_n$ is a character sequence the model is trained to output (eg.  ``knife''). $\phi_1,\phi_2,\ldots,\phi_m$ is a phoneme sequence the model is trained to output (eg. /na\textipa{I}f/), and $\mathcal{L_{\mathbf{x}}}$ is the language identity of the input speech $\mathbf{x}$.}
\label{fig:architecture}%
\end{figure}

In addition to scaling ASR training to 100 languages, a key contribution of our work is the use of a context-independent phoneme objective paired with a language-adversarial classification objective in a
end-to-end grapheme-based neural network, as illustrated in Figure \ref{fig:architecture}.

\subsection{Baseline Model}
%Our models include a Kaldi baseline \cite{povey2011kaldi}: \textbf{describe Kaldi config} and a end-to-end neural model using ESPnet \cite{watanabe2017hybrid}.
%with a similar configuration as used by \newcite{inaguma2018transfer}. (\textbf{Tuning and training time constraints mean we might not use Hirofumi's setup.})
Our experiments are conducted within the framework of a hybrid CTC-attention
end-to-end neural model using ESPnet \cite{watanabe2017hybrid}, which uses an
encoder-decoder architecture implemented in PyTorch \cite{paszke2017automatic}. The encoder we use consists of VGG-like
convolution layers \cite{simonyan2014very,Sercu2016} followed by a multilayer bidirectional long short-term memory (LSTM) \cite{hochreiter1997long, schuster1997bidirectional}. The decoder uses location-based attention \cite{Chorowski2015} and an LSTM.
In addition to the attention, the decoder also incorporates CTC probabilities over graphemes to encourage monotonicity in decoding.

\subsection{Phoneme Objective}

The end-to-end neural model performs direct grapheme prediction without recourse to a pronunciation lexicon as traditional hybrid HMM-DNN models do.
%Since many orthographies mutually disjoint, we use a context-independent phoneme CTC objective to encourage learning of representations that are less dependent on the orthographic idiosyncracies of the languages seen in training.
Since different orthographies may be mutually disjoint or only weakly related to the phonetic content of the input speech, we use a context-independent phoneme CTC objective to encourage  learning of representations independent of such orthographic idiosyncrasies.

%Recent work in hierarchical training objectives supports connecting the phoneme objective an encoder layer before the last layer \cite{krishna2018hierarchical}. In preliminary experiments we found the second last layer to work well.

% In limited preliminary experiments we found using the phoneme objective during adaptation to usually be harmful, so we use it only during the pretraining to encourage language-independence before dropping it in language-specific adaptation.
We performed limited preliminary experiments to determine how best to use the phoneme objective, which corroborated recent work in hierarchical training objectives that supports inserting the phoneme objective in the layers below the final layer \cite{krishna2018hierarchical}. We also found that using the phoneme objective during adaptation was harmful and therefore in all reported experiments we use it only during multilingual pretraining.

%\textbf{Maybe not the case based on some of the 99+phn results.}

\subsection{Language-Adversarial Pretraining}
\label{sec:adv_pretraining}

% For language-adversarial training we used a log-linear 
% classifier over languages seen in pretraining. A mean of the encoder
% states is taken over an utterance, which is then fed into the classifier. This classifier is connected to the second last layer, like the phoneme CTC objective. Between each grapheme/phoneme prediction parameter update, we interleave an adversarial training step for the encoder.
For language-adversarial training we used a log-linear classifier over all languages seen in pretraining. An utterance-level mean of the penultimate encoder layer states is fed into the classifier. For each batch in training we update the network using the interpolated grapheme and phoneme objectives before a separate update step using the adversarial objective.

We follow the learning rate scheduling of \newcite{ganin2016domain},
 where the weight of the adversarial objective relative to the
speech recognition tasks follows $\lambda(p) = \frac{2}{1 + \exp(-10p)} - 1$
 over the course of training, where $p \in [0, 1]$ is a measure of training progress. We drop the adversarial objective during target language adaptation.
% You could keep adversarial training on over readings.

%\subsection{Language models}

%\textbf{Potentially train LMs on the whole Bible text. After cleaning, the audio is just a subset of this. Could also train the LMs just on the subset. It would be interesting to see how character-level universal language models help/hurt models as compared with in-language language models.}

%\subsection{Language models}

%\textbf{Potentially train LMs on the whole Bible text. After cleaning, the audio is just a subset of this. Could also train the LMs just on the subset. It would be interesting to see how character-level universal language models help/hurt models as compared with in-language language models.}

\section{Experimental Setup}

\subsection{Language Versus Reading Adaptation}

We chose as target adaptation languages those languages for which we have multiple
readings of the Bible. This allows us to assess adaptation of the pretrained multilingual model in two scenarios: \emph{language adaptation} and \emph{reading adaptation}. In \emph{reading adaptation}, it is adapted to data from each reading
of the target language, including the reading from which we select held-out evaluation utterances. In \emph{language adaptation} it is adapted only to readings
that are not represented in the evaluation set. This
last case, of adapting to just one or several speakers of a new language (in order to
ultimately have a system that generalizes beyond those speakers in the
language) is not common in speech recognition experimentation. Results and findings for language adaptation will be presented in Section \ref{sec:notgt}.

\begin{table*}[]
    \centering
    \resizebox{\textwidth}{!}{%
    \begin{tabular}{|c|c|cr@{\hspace{0.5ex}}r|cr@{\hspace{0.5ex}}r|cccr@{\hspace{0.5ex}}r|}
    \toprule
     Target & \textsc{Mono} & \multicolumn{3}{c|}{\textsc{Que}} &
	 \multicolumn{3}{c|}{\textsc{Cyr}} &
	 \multicolumn{5}{c|}{\textsc{Que}+\textsc{Cyr}}\\
	 \midrule
		%\cmidrule(rl){3-4} \cmidrule(rl){5-6} \cmidrule(rl){7-10}
       & & - & \multicolumn{2}{c|}{\texttt{+phn+adv}} & - & \multicolumn{2}{c|}{\texttt{+phn+adv}} & - &
	  \texttt{+phn} & \texttt{+adv} & \multicolumn{2}{c|}{\texttt{+phn+adv}}\\
    \midrule
    Aymara & 40.6 & 34.3 & 34.5 & (+0.6\%) &  37.9 & 35.9 & (-5.3\%) & 34.6 & \textbf{34.2} & 34.8 & \textbf{34.2} & (-1.2\%)\\
    SB Quechua & 14.8 & \textbf{13.8} & 14.0 & (+1.4\%) & 16.3 & 17.0 & (+4.3\%) & 14.9 & 14.2 & 14.0 & 13.9 & (-6.7\%)\\
    Indonesian & 14.9 & 15.1 & 15.3 & (+1.3\%) & 16.1 & 17.9 & (+11.2\%) & 15.8 & 15.6 &	15.5 &	\textbf{14.7} & (-7.0\%)\\ 
			   &    &  \multicolumn{2}{|c}{Avg. rel. $\Delta$:} & (+1.1\%) & \multicolumn{2}{r}{Avg. rel.  $\Delta$:} & (+3.4\%) &     & &  \multicolumn{2}{c}{Avg. rel.  $\Delta$:} & (-4.9\%) \\
    %\midrule
    %Aymara-notgt & N/A & 91.4 & \textbf{86.7} & 	89.3 &	87.0 \\
    %SB Quechua-notgt & N/A & 62.3	& 37.2 &	59.1 &	\textbf{34.3} \\
    %Indonesian-notgt & N/A & 24.6	& \textbf{24.5} &	26.2 &	25.0 \\
    \bottomrule
    \end{tabular}
	}
	\caption{Word error rate (\%) comparison of multilingual models adapted to
	target languages, with and without auxiliary training objectives (relative change in parentheses).
	Additionally including Cyrillic-script languages in pretraining (\textsc{Cyr}) doesn't consistently improve over a
	model pretrained on Quechuan languages (\textsc{Que}) unless additional phoneme
	and language-adversarial objectives (\texttt{+phn} and \texttt{+adv}) are
	used in combination (\texttt{+phn+adv}). The auxiliary objectives help when
	pretraining languages are varied, but hinder when they are very
	similar. The final four columns suggest that the objectives are
	complementary. Average relative word error rate change for each pretraining set when adding in the auxiliary objectives (versus no aditional objectives) is indicated by \emph{Avg. rel. $\Delta$}.}
    \label{tab:que_cyr}
    \vspace{-0.3cm}
\end{table*}

\subsection{Training Settings}

We established training, validation and test sets for each reading using a random 80/10/10 split. When pretraining or adapting the multilingual systems, we used the combined training sets of the constituent readings. 

We used 80-dimensional log Mel filterbank features with 3-dimensional pitch features. We tuned hyperparameters for these models using one Aymara reading.\footnote{CMU Wilderness reading ID: AYMSBU.} We found that a 4 layer encoder, 1 layer decoder with 768 for the encoder hidden size and projections, decoder hidden size, and attention hidden size yielded equal-best results with deeper models. These settings were then used for training the models used in our experiments. %\textbf{Tuning results can be found on /export/b13/oadams/espnet-merge3/egs/cmu\_wilderness/asr1/exp}.

For the training objective, we linearly interpolated the the attentional decoder cross-entropy loss with the grapheme CTC and phoneme CTC objectives. Equal weight was given to all three since we found that to be effective in preliminary experiments. Note however, that the effective weight of the adversarial objective effectively changes over the course of training because of the learning rate scheduling mentioned in \S\ref{sec:adv_pretraining}. We trained for 15 epochs in all cases except where otherwise noted.

Note that during adaptation we initialize the model using both the
multilingual encoder and decoder. We found this to work best in preliminary
experimentation on a Spanish reading.

\section{Preliminary Investigation of the Auxiliary Objectives}
\label{sec:pilot}

In this section we evaluate the use of the auxiliary phoneme and
language-adversarial objectives described in Section \ref{sec:aux_objectives}
on
%While in Section \ref{sec:lang_sim} we explore
%more deeply the importance of different types of language relatedness between
%the pretraining and target languages, in this section we present results using
%multilingual pretraining on
two divergent groups of languages that are distinct
along a number of dimensions, including orthography, language family and
phonology, in order to assess the auxiliary objectives' capacity to bridge the
divide between these languages during pretraining. This serves as an initial exploration before further experiments in Section \ref{sec:lang_sim} and Section \ref{sec:notgt}, where we choose from a broader set of pretraining languages.

\paragraph{Pretraining languages} 

We pretrained models on two groups of languages
separately and together. The first consists of six languages from the Quechuan
language family, including sub-varieties of Quechua I and II (qub, quf, qvs, qvw, qwh and qvh). We henceforth refer to this group as
\textsc{Que}. The second consists of six languages that use the Cyrillic
script and we refer to this group as \textsc{Cyr}. These languages include
Nogai (nog), Bashkir (bak), Gagauz (gag), Khakas (kjh), Crimean Tatar (crh),
and Russian (rus). With the exception of Russian, these languages are all
Turkic. The character sets do not overlap between \textsc{Que} and \textsc{Cyr} and this was a deliberate choice in this preliminary experiment to maximize the differences between the two groups.

\paragraph{Evaluation languages} To test the pretrained models in varied
contexts, we evaluate our models on three languages: Central Aymara (ayr),
South Bolivian Quechua (SB Quechua; quh), and Indonesian (ind). These languages vary in a
number of dimensions: SB Quechua is very closely related to
\textsc{Que}, while Indonesian is distant; Aymara is phonologically very
similar to Quechuan languages, but is considered to be from a different family;
Aymara had a high monolingual baseline error rate, while the others are lower;
and Indonesian has three readings while the others have two. However, all evaluation
languages use the Latin script. Note that in this section we assess
performance in the reading adaptation case, while Section \ref{sec:notgt} presents results on the held-out reading case.

\begin{figure*}
\begin{center}
\includegraphics[width=8cm]{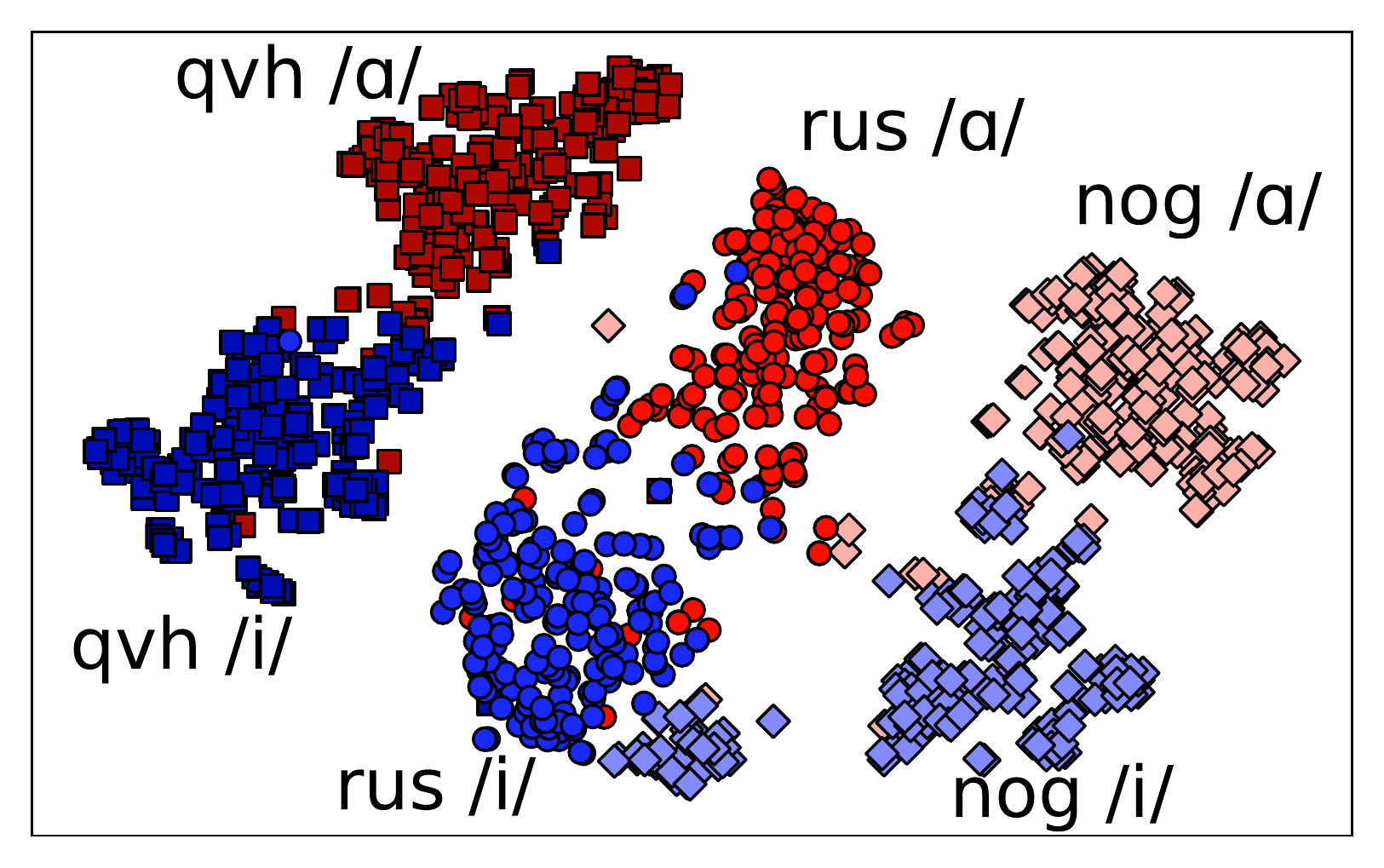}
\hspace{-0.2cm}
\includegraphics[width=8cm]{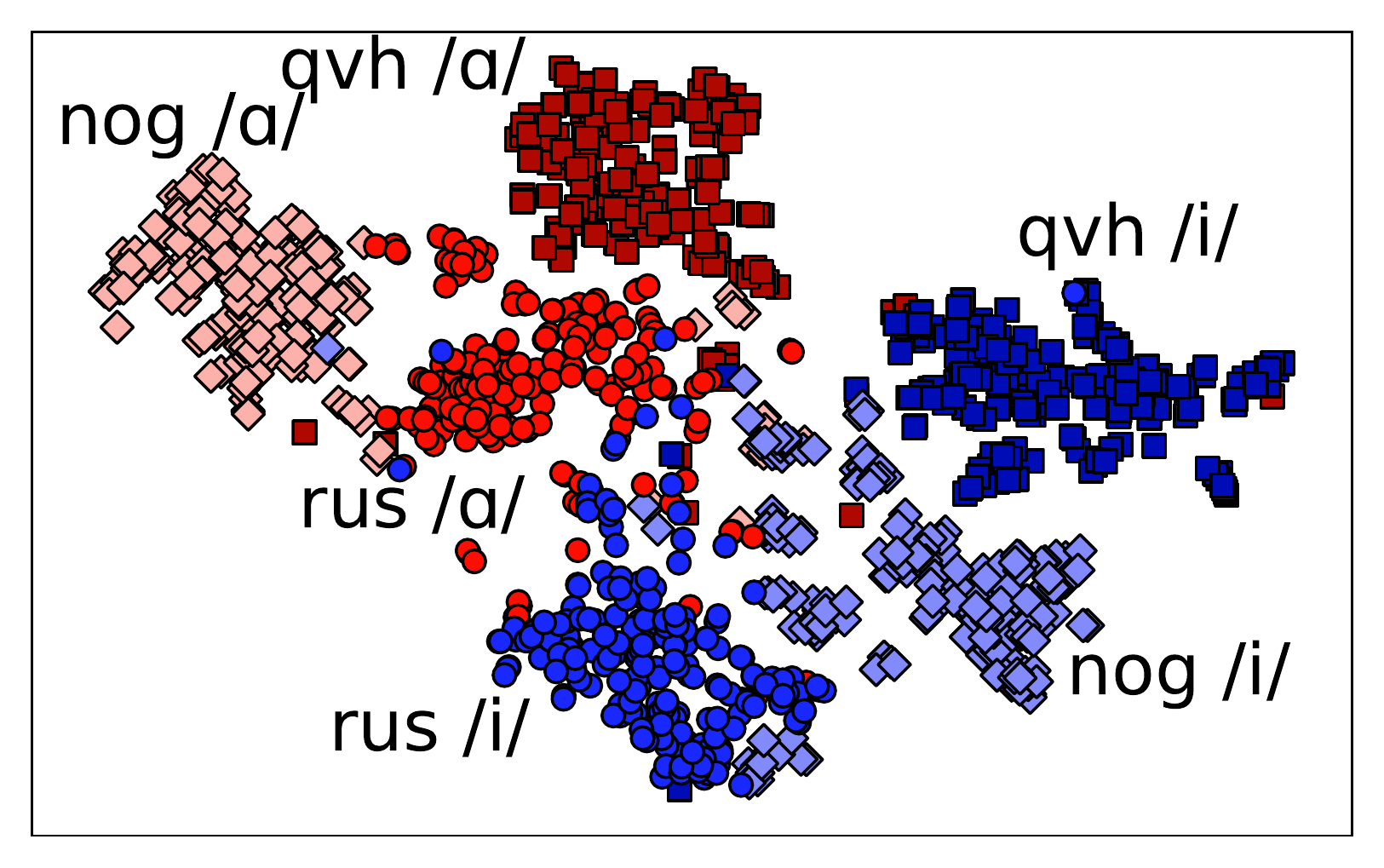}
\end{center}
\vspace{-0.3cm}
\begin{center}
\includegraphics[width=7cm]{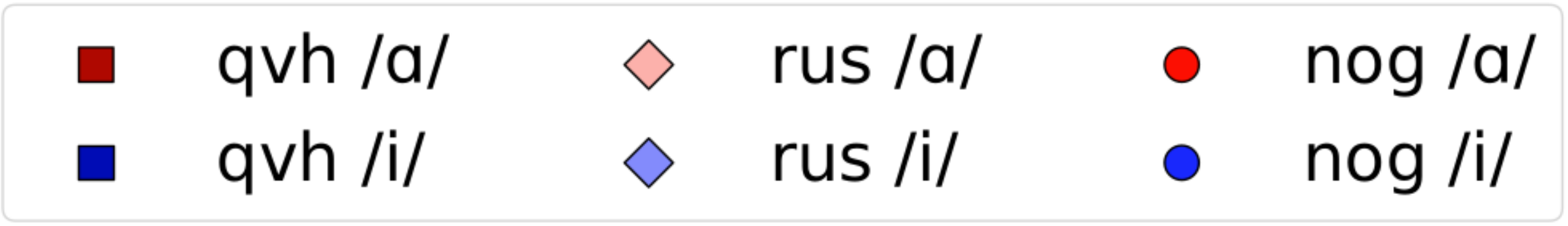}
\end{center}

\caption{t-SNE representation of encoder states corresponding to /\textipa{A}/ and
/i/ across Quechua (Huamalies Dos de Mayo; qvh), Russian (rus), and Nogai (nog). \emph{Left}: the model
without the phoneme and adversarial objective. \emph{Right}: the phoneme and language-adversarial objectives are added in, causing phoneme clusters between languages to
gather closer together, and language to become less relevant in cluster placement.}
\label{fig:tsne}
\vspace{-0.3cm}
\end{figure*}

\paragraph{Experiments}
%We first examine pretraining on both language groups, which we will call \textsc{Que+Cyr}.
Table \ref{tab:que_cyr} compares the performance of monolingual target-language models to models adapted to the target language after being pretrained on \textsc{Que}, \textsc{Cyr} and their combination, \textsc{Que+Cyr}.
\textsc{Cyr} pretraining underperforms pretraining with \textsc{Que}
for all evaluation languages likely due to the orthographic mismatch with all of the evaluation languages. The model
pretrained on \textsc{Que+Cyr} also underperforms \textsc{Que}.
Introducing the auxiliary phoneme and language-adversarial objectives helps
to overcome this performance loss, making the
\textsc{Que+Cyr}-pretrained model the best for adaptation to Aymara and Indonesian.
\textsc{Que} remained the best pretraining set for adaptation to SB
Quechua, which is unsurprising given how well represented SB Quechua is by the
languages included in the Quechuan language group. This suggests that when a
substantial amount of data in very closely related languages is available (in
this case, close to 100 hours of \textsc{Que} data), then there is little
to be gained from highly unrelated languages.

When pretraining on \textsc{Que} and \textsc{Cyr} separately, the
auxiliary objectives underperformed baseline multilingual pretraining on
average. The variation in languages within these groups is far less than the
variation between groups. Given that the phoneme and adversarial objectives
are intended to overcome variation between pretraining languages, this result
indicates that there must be a sufficient level of diversity in the pretraining
languages before the auxiliary objectives are of benefit when adapting to
certain target languages.

Results from pretraining on \textsc{Que+Cyr} showed either objective
to help on average, and that the effects are complementary. Because of this, we
opted to include them together in subsequent experimentation. We evaluated this
best performing model on the larger set of other evaluation languages.
Results in Table \ref{tab:phon_inv_geo_ninetynine} show that in all cases multilingual
pretraining of \textsc{Que+Cyr} with the auxiliary objectives outperformed its counterpart without
the objectives (which frequently undeperformed the monolingual model), and in
all but one case this led to an improvement over the monolingual baseline.\footnote{However, this doesn't hold in the \emph{language adaptation} scenario, where the auxiliary objectives help \textsc{Que+Cyr} only slightly; see Section \ref{sec:notgt}.}

To gain insight into how the auxiliary objectives change the representation
of speech learnt by the models, we applied 2D t-SNE dimensionality
reduction \cite{vandermaaten2008tsne}. Figure \ref{fig:tsne} plots the
representations of two phonemes in three languages learnt by the encoder\footnote{We
established the correspondence between encoder states and phonemes by using forced alignment with Kaldi
\cite{povey2011kaldi}, taking the encoder state at the mid-point of the
duration on the phonemes.} in the case without and with the auxiliary objectives.
In the multilingual pretraining baseline, six clusters are
represented for each language--phoneme combination. These appear stratified by
language, with different phoneme clusters within languages close to one
another. With the auxiliary objectives, phoneme clusters between languages move
closer to one another, while language identity becomes less relevant in determining which
phoneme clusters neighbour one another. In the latter plot, the Nogai phonemes become  separated by a Russian /\textipa{A}/. This is particularly salient since the Nogai speaker was female, while the Russian speaker had a deep male voice.

\section{Reading Adaptation}
\label{sec:lang_sim}
%\textbf{Renamed section title to add ``reading adaptation'' qualification, but I'm not sure.}

\begin{table*}[]
    \centering
    \resizebox{\textwidth}{!}{%
    \begin{tabular}{|l|c|cr@{\hspace{0.5ex}}r|cr@{\hspace{0.5ex}}r|cr@{\hspace{0.5ex}}r|cr@{\hspace{0.5ex}}r|}
    \toprule
    & \textsc{Mono} & \multicolumn{3}{c|}{\textsc{Que+Cyr}} & \multicolumn{3}{c|}{\textsc{Phonology}} & \multicolumn{3}{c|}{\textsc{Geo}} &
	\multicolumn{3}{c|}{\textsc{100-lang}} \\
	%\cmidrule(rl){3-4} \cmidrule(rl){5-6} \cmidrule(rl){7-8} \cmidrule(rl){9-10}
	\midrule
    & & - & \multicolumn{2}{c|}{\texttt{+phn+adv}} & - & \multicolumn{2}{c|}{\texttt{+phn+adv}} & - & \multicolumn{2}{c|}{\texttt{+phn+adv}} & - & \multicolumn{2}{c|}{\texttt{+phn+adv}} \\
    \midrule
    ayr & 40.6 & 34.6 &	34.2 & (-1.2\%)& \textbf{33.9} & 34.5 & (+1.8\%) &  35.4   &     34.9 & (-1.4\%)           & 34.2 & 34.5 & (+0.9\%) \\
    quh & 14.8 & 14.9 &	\textbf{13.9} & (-6.7\%)& 14.4 & 14.5 & (+0.7\%) & 15.5 &  14.8 & (-4.5\%) & 15.1 & 14.7 & (-2.6\%) \\
    kek & 23.9 & 24.8 &	23.7 & (-4.4\%)& 24.8 & 24.5 & (-1.2\%) & 23.0 &  \textbf{22.9} & (-0.4\%) & 24.9 & 24.4 & (-2.0\%) \\
    ixl & 20.7 & 21.2 &	20.1 & (-5.2\%)& - & \multicolumn{2}{c|}{-} & \textbf{19.7} & 20.1 & (+2.0\%)                                & 20.8 & 20.6 & (-1.0\%) \\
    mlg & 45.2 & 43.5 &	\textbf{41.4} & (-4.8\%)& 43.2 & 41.7 & (-3.5\%) &  43.3    &  42.2 & (-2.5\%)         & 44.4 & 42.2 & (-5.0\%) \\
    ind & 14.9 & 15.8 &	14.7 & (-7.0\%)& \textbf{13.7} & 14.3 & (+4.4\%) & 14.0 &  \textbf{13.7} & (-2.1\%) & 14.7    & 14.2 & (-3.4\%) \\
    kia & 14.6 & 14.6 &	13.2 & (-9.6\%)& - & \multicolumn{2}{c|}{-} & \textbf{12.1} & \textbf{12.1} & (-0.0\%)                                & 14.4 & 13.0 & (-9.7\%) \\
    swe & \textbf{20.5} & 22.7 & 21.6 & (-4.9\%) & 26.4 & 24.2 & (-8.3\%) & 22.0 &  21.2 & (-3.6\%) & 23.9 & 24.6 & (+2.9\%) \\
	spn & 14.5 &	19.7 & 14.4 & (-26.9\%) & 13.9 & 13.8 & (-0.7\%)  & 13.1 &    \textbf{12.1} & (-7.6\%)                    & 15.8 & 14.8 & (-6.3\%)  \\
	%\emph{avg} & \emph{23.3} & \emph{23.5} & \emph{21.9} (-7.9\%) & \emph{24.3} & \emph{23.9} & \emph{22.0} & \emph{\textbf{21.6}} & \emph{23.1} & \emph{22.6} \\
	&  & \multicolumn{2}{r}{Avg. rel. $\Delta$:} & (-7.8\%)   &  \multicolumn{2}{r}{Avg. rel. $\Delta$:} & (-1.0\%) & \multicolumn{2}{r}{Avg. rel.  $\Delta$:} & (-2.3\%) &  \multicolumn{2}{r}{Avg. rel.  $\Delta$:} & (-2.9\%) \\
%    \midrule                #VALUE                            
%    ayr-notgt \\
%    quh-notgt & 62.3 & 37.0 & 43.9 & \textbf{x} & 42.8 \\
%    kek-notgt & 75.6 & 74.1 & 74.1 & \textbf{73.8} & 74.4\\
%    mlg-notgt \\
%    ind-notgt & 24.6 & 21.5 & 20.8 \\
%    swe-notgt & 72.9 & 79.4 & 63.2 & 75.4 & \textbf{62.5}\\ 
%    spn-notgt \\
    \bottomrule
    \end{tabular}
	}
    \caption{Word error rate (\%) comparison of adaptation of models pretrained on: Quechuan and Cyrillic-script languages (\textsc{Que+Cyr}), languages phonologically and phonetically similar to the target (\textsc{Phon/Inv}),
	geographically proximate languages (\textsc{Geo}), and a massively multilingual set of languages (\textsc{100-lang}). In each case we compared the average relative WER change when adding  	 auxiliary phoneme and language-adversarial objectives (\texttt{+phn+adv}). Dashed entries had
	phonology and phonetic inventories that weren't well attested in URIEL, so were not
	assessed.}
    \label{tab:phon_inv_geo_ninetynine}
\end{table*}

In the previous section we explored the use of two dissimilar groups of
languages in a multilingual setup. Multilingual pretraining of languages from a
different language family and script benefitted from an explicit phoneme
objective and adversarial objective when there was sufficient
diversity in the pretraining languages. However, a
change in orthography was conflated with a change in language family,
geographic location, and phonological/phonetic characteristics. In this section, we investigate which factors are most important in choosing languages for multilingual pretraining and how useful it is to scale up model pretraining to many languages. This exploration is conducted in the reading adaptation scenario; language adaptation with unseen target speakers is addressed in Section \ref{sec:notgt}.
%Does this mean phonetic/phonological
%similarity, geographic proximity, or relatedness by language family?
Beyond answering these questions, this investigation
reveals more information about the utility of the proposed auxiliary
objectives in different scenarios.

\paragraph{Phonology \& Geography}

We test across a number of evaluation languages (c.f. Table
\ref{tab:eval_stats}) by determining, for each evaluation language, groups of
pretraining languages that are similar to the evaluation languages in different
ways. In order to determine language similarity in a principled way we used URIEL and \emph{lang2vec}
\cite{Littell2017} to produce feature vectors for each language based on
information from several linguistic resources before
calculating their cosine similarity. For each language we used two feature
vectors. The first is a concatenation of the lang2vec
\texttt{phonology\_average} and \texttt{inventory\_average}
vectors, characterizing phonological properties and phonetic inventory. The second
represents geographic location. We denote these two groups
\textsc{Phon/Inv} and \textsc{Geo} respectively.\footnote{We didn't create \textsc{Phon/Inv} sets for Ixil
and Garap because their phonological features and phonetic inventories were not
well attested, and we didn't use the lang2vec language family
vectors since most of the Quechuan languages were not
captured as being highly similar to SB Quechua.} Geographic proximity may serve as a proxy for other similarities not captured in \textsc{Phon/Inv}, including language family, orthographic similarity, and the likelihood of exchanged loan words.

We filtered for languages in the dataset
with \texttt{good} or \texttt{very good} alignments before ranking them by
cosine similarity with the evaluation languages
in terms of phonological and phonetic similarity as well as geographical
proximity. To create each of the pretraining sets, we took between 7 and 14 of the top languages, matching
approximately the total duration of the phonetically/phonologically similar
groups with the geographically proximate language groups.\footnote{An exhaustive list of the CMU Wilderness language codes for each pretraining group can be found in Appendix \ref{sec:lang_lists}, along with durations of each pretraining set.} For most languages, there is no overlap between the \textsc{Geo} and \textsc{Phon/Inv} sets.

\paragraph{Massively multilingual model} As a further point of comparison, we
pretrain a model on around 100 languages (denoted \textsc{100-lang}), for approximately 1650 training hours in total.\footnote{These models were pretrained for 6 epochs.}

%\textbf{At 400 or so hours, the South american verygood model will give a sense of how useful the adversarial model is as data scales. It's about a logarithmic halfway point to the full data.}

\subsection{Auxiliary Objectives Findings}

The results in Table \ref{tab:phon_inv_geo_ninetynine} extend on our findings
in Section \ref{sec:pilot}, continuing to support the benefit of the use of the auxiliary
objectives while shedding more light on the type of language variability the
objectives help to overcome. \textsc{Geo} and \textsc{100-lang} benefitted
comparably from the objectives on average, while \textsc{Phon/Inv} did less so. \textsc{Que+Cyr} benefitted the most. This suggests that the objectives may help more when pretraining languages are orthographically, phonetically and phonologically diverse.

Unlike the other languages, the Swedish
\textsc{Phon/Inv} vectors were not well attested. As a result the Swedish
\textsc{Phon/Inv} group has languages with a similar phonetic inventory that
were also unattested phonologically. This model underperformed the
monolingual model by a large margin, suggesting that similarity of phonetic
inventory alone may not be so useful alone without similarity of phonological features. Models pretrained on this set also benefitted the most from the
auxiliary objectives. It may be the case that the auxiliary
objectives push together representations of allophones within languages, and
pronunciation variations of the same phonemes between languages. When Swedish
is discounted, the average relative improvement when adding auxiliary objectives for
\textsc{Phon/Inv} becomes negligable.

The \textsc{Phon/Inv} configurations are hurt by the auxiliary objectives for SB Quechua and Aymara and Indonesian. The \textsc{Phon/Inv} sets for the first two of these languages emphasized Quechuan languages, and this
corroborates the indication in Section \ref{sec:pilot} that the auxiliary
objectives may not help so much when pretraining languages are similar. On the other hand, the Indonesian \textsc{Phon/Inv} included Afro-Asiatic and Niger-Congo languages, as well an Indo-European language and Huave, a language isolate from Mexico, yet it was not improved by auxiliary objectives.
%However,
%as a counter-point Indonesian \textsc{Phon/Inv} was hurt by them, despite
%geographic and language family variation of its member langs.

%\item the mlg-Phon/Inv set was geographically and lang-fam diverse. swe-Phon/Inv was largely niger-congo, and far from sweden. both of these langs got the best help from the adversarial objective, so corroborating earlier results adv+phn seems to help diverse models the most.

%\item basically comparing (1) +phn+adv to their respective other models in
%all cases; (2) geo to Phon/Inv, and 99-lang with mono.

%\textbf{Just realized Gagauz is included twice, with two different scripts, in
%the Spn corpus.}

\subsection{Choice of Pretraining Languages}

The average relative word error rate (WER) change for \textsc{Geo} against
\textsc{Phon/Inv} was -2.2\% without auxiliary objectives, and -4.4\% with
them,\footnote{Discounting Swedish, this becomes +0.2\% and -3.1\%.} suggesting
that features correlated with geography are useful for guiding pretraining
language selection. Counter-examples were Aymara, SB
Quechua and Malagasy, which performed worse when pretrained on \textsc{Geo}. In the case of SB Quechua, only one Quechuan language was represented in \textsc{Geo} (Inga), while \textsc{Phon/Inv} had three (qub, qvh, quf). Madagascar is far
removed from where most Austronesian languages are spoken, so Malagasy's
\textsc{Geo} set were almost all Niger-Congo languages, while the
\textsc{Phon/Inv} had a diverse array of Austronesian, Indo European,
Afro-Asiatic, Sino-Tibetan and Mayan languages. However, on average, these results
suggest that geographical proximity is a decent guide to pretraining language selection. Another advantage is that it requires no explicit phonological features, making it applicable to a much larger number of languages.

%\item spn-geo+phn+adv had the biggest gain over mono, and probably also over
%sim-Phon/Inv. spn-geo included a wide range of languages, indo european,
%turkic, afro-asiatic, niger-congo

The average relative WER change of \textsc{100-lang} against \textsc{Mono} was
+1.3\%, indicating that massively multilingual pretraining by itself not useful
if the target speakers are seen in training. Using the auxiliary objectives overcame the difference, resulting in a -1.6\% average relative WER change. However, pretraining with \textsc{Geo}\texttt{+phn+adv} yielded an average relative delta of -7.4\% over the monolingual model. Though more languages help, they are not
necessarily better than geographically proximal languages (however, results are very different when not adapting to target speakers: see Section \ref{sec:notgt}).
%\item swe gets closer lang fam relatedness from geo. kek-geo is almost
%all mayan, which gets a big gain (kek-Phon/Inv had 4 mayan languagse,
%but ones from other families).

%\item We trained an Austronesian model on the languages with very good
%alignment. 43.0 for mlg and 14.4 for ind, so didn't get gains despite
%high quality corpora. geographic proximity seems to matter more, though
%it also indicates lang family often. ind-geo was almost all
%austronesian but more closely related. mlg-geo was almost all
%niger-congo. but mlg+phon+adv was very diverse by lang family and
%performed the best.

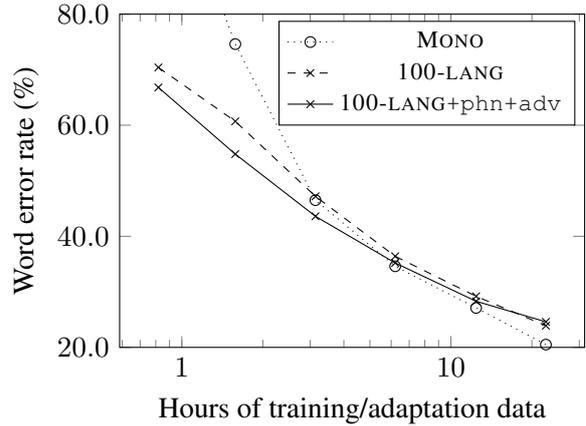
\begin{figure}
\begin{tikzpicture}
\begin{axis}[
	xlabel=Hours of training/adaptation data,
	ylabel=Word error rate (\%),
	width=\columnwidth,
	xmode=log,
	ymax=80,
	ymin=20,
    height=6cm,
    legend style={font=\small},
	log ticks with fixed point,
	yticklabel style={
			/pgf/number format/fixed,
			/pgf/number format/precision=1,
			/pgf/number format/fixed zerofill
	},
	xticklabel style={
			/pgf/number format/fixed,
			/pgf/number format/precision=0,
			/pgf/number format/fixed zerofill
	}
	]

\addplot[black, dotted, mark=o, mark options={solid}]
coordinates {
	(0.818289, 112.2)
	(1.58023, 74.6)
	(3.13732, 46.5)
	(6.20865, 34.6)
	(12.407,  27.1)
	(22.5402, 20.5)
};

\addplot[color=black, dashed, mark=x, mark options={solid}] 
coordinates{
	(0.818289, 70.4)
	(1.58023, 60.7)
	(3.13732, 47.2)
	(6.20865, 36.4)
	(12.407, 29.2)
	(22.5402, 23.9)
};

\addplot[black, solid, mark=x]
coordinates {
	(0.818289, 66.8)
	(1.58023, 54.8)
	(3.13732, 43.6)
	(6.20865, 35.2)
	(12.407, 28.3)
	(22.5402, 24.6)
};
\legend{\textsc{Mono},\textsc{100-lang},\textsc{100-lang}\texttt{+phn+adv}}
\end{axis}
\end{tikzpicture}
\caption{Scaling training/adaptation data for Swedish. Adapting to the full
dataset, the auxiliary objectives underperformed both the monolingual and baselines, but yields an advantage when the model is adapted to less target language data.}
\label{fig:swedish_scaling}
\end{figure}

In two cases pretraining with \textsc{100-lang} was hindered
by the auxiliary objective. In one of these cases, Swedish, both
\textsc{100-lang} variations substantially underperformed the monolingual
baseline. One possible reason is that there is enough target language and
speaker data that the multilingual pretraining and auxiliary objectives offer no benefit. We scaled
training/adaptation data for Swedish from under 1 hour. Figure
\ref{fig:swedish_scaling} indicates that in this case the
auxiliary objectives do lead to better initialization, with gains
being lost only when around 5 hours of target language and reading data are seen.

\section{Language Adaptation}
\label{sec:notgt}
Previous sections have addressed the reading adaptation scenario, where the ASR model is adapted to speech from the target reading (ie. where target speakers have been heard in adaptation). In this section we evaluate in a language adaptation scenario, adapting to readings in the target language, but not the target reading. The question of how well a multilingual model can be
adapted to a language on the basis of recordings from a small number of target-language
speakers is relevant to incident response situations such as those
modelled by LORELEI \cite{Strassel2016}, where a single language consultant
is available for which recorded speech can be made. We performed
experiments analogous to those of the previous sections where the evaluation reading was not seen in training or
adaptation. This is a challenging task as the model must generalize to multiple
speakers of a language on the basis of seeing only several in training.
Most of the findings corroborate what was found in the previous sections. Here we highlight differences.

\paragraph{Massively multilingual pretraining} led to substantially better performance than other methods, unlike in the reading adaptation task. For each evaluation language, the \textsc{100-lang} model outperformed the next best method, with one exception: Indonesian. In that case \textsc{Geo} set performed the best, as the languages were not only geographically proximate, but also consisted
    entirely of other Austronesian languages. The takeaway (c.f. Table \ref{tab:notgt}) is that you should always use more pretraining languages unless you know your target speakers, as in the reading adaptation scenario.

\paragraph{Auxiliary objectives} remained useful on the whole. However, while the difference in WER achieved when
adding the auxiliary objectives was similar to those reported in Section \ref{sec:lang_sim} for
\textsc{Phon/Inv} and \textsc{100-lang}, \textsc{Geo} and \textsc{Que+Cyr} no longer achieved improvements. \textsc{Que+Cyr} notably only achieved a -0.2\% average relative WER
change when adding the auxiliary objectives, while achieving -7.8\% in the reading adaptation case.  While the auxiliary objectives remained useful on the whole, their effect was dwarfed by the value of adding more languages.

\paragraph{Phonology versus Geography} \textsc{Geo} sets with or without auxiliary objectives lost their edge over \textsc{Phon/Inv}, with high
variance in scores. The amount of training data becomes the
dominating variable affecting WER.

\begin{table}[]
    \centering
    \resizebox{\columnwidth}{!}{%
    \begin{tabular}{cccccr@{\hspace{0.5ex}}r}
    \toprule
         & \textsc{Mono} &
		 \begin{tabular}{@{}c@{}}\textsc{Que}\\+\textsc{Cyr}\end{tabular}
		 & \begin{tabular}{@{}c@{}}\textsc{Phon}\\+\textsc{Inv}\end{tabular} & \textsc{Geo} & \multicolumn{2}{c}{\textsc{100-lang}}\\
    \midrule
    ayr  & 91.4 & 86.3    & 86.7 & 87.2 & \textbf{79.2} & (-8.2\%) \\
    quh  & 62.3	& 35.8    & 35.5 & 42.8 & \textbf{30.1} & (-15.2\%)\\
    kek  & 75.6	& 74.3    & 73.8 & 74.4 & \textbf{73.5} & (-0.4\%)  \\
    ixl  & 81.8	& 79.8    &  -  &  78.4   & \textbf{74.3} &  (-6.9\%)  \\
    mlg  & 103.6& 68.3    & 64.0 & 63.7 & \textbf{62.2} & (-2.4\%)  \\
    ind  & 24.6	& 23.5    & 22.1 & \textbf{21.1} & 21.6 & (+2.4\%)   \\
    kia  & 57.2	& 51.5    &   -   &  49.9  & \textbf{48.2} & (-6.4\%)  \\
    swe  & 72.9	& 64.4    & 75.4 & 62.5 & \textbf{55.1} & (-11.8\%)\\
    spn  & 44.8	& 33.8    & 33.4 & 32.7 & \textbf{29.9} & (-8.6\%) \\
    \multicolumn{6}{r}{Avg. rel. $\Delta$ of \textsc{100-lang} wrt. next best method:} & (-6.0\%)\\
%	  & &  \multicolumn{4}{c}{Average relative $\Delta$ of \textsc{100-lang} wrt. next best method: -6.2\%}\\
    %& & & & & \multicolumn{155.8}{c54.9}{Avg $\delta$: -2.5\%}\\
    \bottomrule
    \end{tabular}
	}
    \caption{Adaptation to the non-target reading in the target language. All
	language sets use the auxiliary training objectives, which again exhibited an
	relative gain over the corresponding model without. The relative deltas of \textsc{100-lang} are with respect to the next closest model on a language-by-language basis.}
    \label{tab:notgt}
\end{table}

\section{Conclusions}

We have explored the utility of pretraining multilingual models on a
variety of language sets, scaling to as as many as 100 languages. Our
experiments have demonstrated the value of auxiliary phoneme and
language-adversarial pretraining objectives in a multilingual end-to-end ASR
framework, particularly when the pretraining languages are diverse. Our results
suggest how to pick pretraining languages when target speakers are seen in the adaptation data: find geographically
proximal languages. When adapting to just several non-target speakers, exposure to more speech in pretraining is the most important thing for model generality, even
if from a wide range of dissimilar languages.

\section*{Acknowledgments}

We would like to thank Tim Baldwin for an off-hand comment that planted the language-adversarial idea in the first author's head, and to Trevor Cohn for some related discussion. Thanks also go to Alexis Michaud and the reviewers for comments.
%\end{itemize}

%The acknowledgments should go immediately before the references.  Do
%not number the acknowledgments section. Do not include this section
%4when submitting your paper for review. \\

\bibliography{library}
\bibliographystyle{acl_natbib}

\appendix

%\section{Appendix A: Full language list}
%\label{sec:eval_lang_list}
%Below is a table of the names, iso-639-3 codes, and CMU Wilderness language IDs of chosen evaluation readings for each evaluation language.
%    \resizebox{\columnwidth}{!}{%
%    \begin{tabular}{lll}
%		\toprule
%		Name & iso 639-3 & ID\\
%		\midrule
%		Aymara (Central) & ayr & AYMSBU\\
%		Quechua (South Bolivian) & quh & QUHRBV\\
%		Kekchi & kek & KEKIBS\\
%		Ixil & ixl & IXIWBT\\
%		Malagasy & mlg & MLGRPV\\
%		Indonesian & ind & INZSHL\\
%		Garap & kia & KIAWBT\\
%		Swedish & swe & SWESFV\\
%		Spanish & spn & SPNBDA\\
%		\bottomrule
%	\end{tabular}
%	}

%\section{Appendix B: Similar languages}
\section{List of readings in each language set}
\label{sec:lang_lists}

Below is a collection of lists of the CMU Wilderness reading codes that comprise different groupings. This includes the target language readings; the Quechuan group; the Cyrillic-script group; the phonologically similar and geographically similar sets for each target language; and the massively multilingual set.

\paragraph{Target language readings}
MLGEIV,
MLGRCV,
MLGRPV,
IX1WBT,
IXIWBT,
IXLWBT,
INZNTV,
INZSHL,
INZTSI,
QUHRBV,
QUHSBB,
QEJLLB,
QUBPBS,
QUFLLB,
QVSTBL,
QVWTBL,
QWHLLB,
SPNBDA,
SPNWTC,
KIABSC,
KIAWBT,
KEKIBS,
KEKSBG,
SWESFB,
SWESFV,
AYMSBU,
AYMBSB.

\paragraph{Evaluation readings}
AYMSBU,
MLGRPV,
IXIWBT,
INZSHL,
QUHRBV,
SPNBDA,
KIAWBT,
KEKIBS,
SWESFV.

\paragraph{\textsc{Que} (97.6 training hours)}
QEJLLB,
QUBPBS,
QUFLLB,
QVSTBL,
QVWTBL,
QWHLLB.

\paragraph{\textsc{Cyr} (59.6 training hours)}
NOGIBT,
BAKIBT,
GAGIB1,
KJHIBT,
RUSS76,
CRHIBT.

\paragraph{\textsc{ayr-\textsc{Phon/Inv}} (145.3 training hours)}
QUBPBS,
TOBBSA,
QUFLLB,
QVSTBL,
INBWBT,
QEJLLB,
JICWBT,
QU1LSM,
QUTIBS.

\paragraph{\textsc{ayr-\textsc{Geo}} (146.2 training hours)}
IGNSBB,
TNATBL,
GNWNTM,
ESENTM,
MCBTBL,
GYRSBB,
CBSBSP

\paragraph{\textsc{quh-\textsc{Phon/Inv}} (177.9 training hours)}
TOBBSA,
DUGBTL,
QUBPBS,
TZHSBM,
HUSLLB,
NYFBTL,
NCUWBT,
QEJLLB,
QUFLLB,
HAGGIL,
NZIBSG,
MNBTBL.

\paragraph{\textsc{quh-\textsc{Geo}} (178.5 training hours)}
GNWNTM,
IGNSBB,
TOBBSA,
ENXBSP,
GYRSBB,
CAXSBB,
CEGNTP,
TNATBL,
ESENTM,
TERTBL.

\paragraph{\textsc{kek-\textsc{Phon+Inv}} (142.1 training hours)}
QU1LSM,
QUTIBS,
TZTWBT,
TUFWYI,
QWHLLB,
PAGPBS,
UDUSIM,
YUASBM.

\paragraph{\textsc{kek-\textsc{Geo}} (137.0 training hours)}
MOPWBT,
POHBSG,
CA1WBT,
CKIWBT,
TZTWBT,
QU1LSM,
QUTIBS,
BZJBSW.

\paragraph{\textsc{mlg-\textsc{Phon/Inv}} (198.2 training hours)}
RONBSR,
TGLPBS,
KVNWBT,
HUVTBL,
KBRSIM,
TPMWBT,
BTXLAI,
KACUBS,
WMWWYI,
IGNSBB,
HAEBSE,
IBATIV,
HILHPV,
TZBSBM.

\paragraph{\textsc{mlg-\textsc{Geo}} (205.38 training hours)}
WMWWYI,
VMWBSM,
MFEBSM,
SEHBSM,
TOHSBM,
CCESBM,
KDCPBT,
CWEPBT,
KKIBST,
NYYBST,
KSBBST,
KDNBSZ,
DUGBTL,
GOGBST.

\paragraph{\textsc{ind-\textsc{Phon/Inv}} (193.1 training hours)}
IBATIV,
TGLPBS,
HAEBSE,
KERABT,
KACUBS,
NYFBTL,
RONBSR,
CWTATB,
HUVTBL,
BTXLAI,
IGNSBB,
JAVNRF,
DUGBTL,
MNKBSG.

\paragraph{\textsc{ind-\textsc{Geo}} (191.5 training hours)}
SUNIBS,
NIJLAI,
JAVNRF,
PSELAI,
IBATIV,
PTULAI,
MVPLAI,
PPKLAI,
BEPLAI,
NPYLAI,
LEWLAI,
MWVLAI.

\paragraph{\textsc{swe-\textsc{Phon/inv}} (122.4 training hours)}
KDJBSU,
NZIBSG,
ANVWBT,
DGABSG,
SHKBSS,
SLDTBL,
KUSTBL,
MUYWBT,
NCUWBT,
LIABSL,
CKOGIL.

\paragraph{\textsc{swe-\textsc{Geo}} (122.4 training hours)}
RMCWFW,
EN1NIV,
RMORAM,
RONBSR,
GAGIB1,
GAGIBT,
CRHIBT,
KPVIBT,
LTNNVV,
ALSBSA,
UDMIBT,
XALIBT,
BAKIBT.

\paragraph{\textsc{spn-\textsc{Phon/Inv}} (123.7 training hours)}
KVNWBT,
HAEBSE,
HUVTBL,
GUGRPV,
HUSLLB,
GUMTBL,
NYFBTL,
KWIWBT.

\paragraph{\textsc{spn-\textsc{Geo}} (129.5 training hours)}
PORARA,
LTNNVV,
EN1NIV,
RMORAM,
ALSBSA,
RMCWFW,
RONBSR,
GAGIB1,
GAGIBT,
CRHIBT,
TAQWBT,
FUQWBT,
MYKWBT.

\paragraph{\textsc{100-lang} (1646.8 training hours)}
OBOWBT,
ACUTBL,
SEYWBT,
HAUCLV,
BZHPNG,
AMKWBT,
GAGIB1,
GNWNTM,
URBWBT,
RUGWBT,
PAUUBS,
SEHBSM,
SNNWBT,
KQETBL,
TGOTBL,
NOGIBT,
XTMTBL,
OJ1CBS,
TNATBL,
AIAWYI,
PABTBL,
MEJTBL,
TWBOMF,
HUSLLB,
ESENTM,
BAKIBT,
HNNOMF,
IFAWBT,
ENXBSP,
ALJOMF,
PXMBSM,
JAISBG,
PIRWBT,
DOMBEC,
NINWYI,
BEPLAI,
JAMBSW,
TERTBL,
LAWNTM,
URATBL,
AGNWPS,
TPIPNG,
TTCWBT,
HUUTBL,
NPYLAI,
KJHIBT,
AZZTBL,
COKWBT,
KWIWBT,
SABWBT,
PADTBL,
GUMTBL,
CRHIBT,
QXRBSE,
RMORAM,
NHYTBL,
TPPTBL,
TUFWYI,
ZLMAVB,
PRFWBT,
TWULAI,
GAGIBT,
FARWBT,
OM1TBL,
RUSS76,
PTULAI,
MIFWBT,
MIYWYI,
MRWNVS,
KNETBL,
PBCBSS,
MYYWBT,
ACHBSU,
ACNBSM,
ADETBL,
AHKTBS,
AK1BSG,
ALPWBT,
ALSBSA,
ALTIBT,
ANVWBT,
ATGWYI,
AVNWBT,
AVUWBT,
AYMBSB,
AYMSBU,
AZEBSA,
BEXWBT,
BQJATB,
BTXLAI,
BZJBSW,
CA1WBT,
CARBSS,
CAXSBB,
CBSBSP,
CMRWBT,
CNLTBL,
CNMRGB,
CRNWBT.
\end{document}